\documentclass[runningheads]{llncs}

\usepackage{times}
\usepackage{latexsym}
\usepackage{pifont}
\usepackage{xcolor}

\usepackage{graphicx}
\usepackage{amsmath,amssymb} 
\usepackage{mathtools}
\usepackage{color}
\usepackage{booktabs}
\usepackage{algorithm}
\usepackage{algpseudocode}
\usepackage[export]{adjustbox} 
\usepackage{array,multirow}
\usepackage{url}
\usepackage{tikz}

\newcommand{\xmark}{\ding{53}}
\begin{document}
\title{Multi-Page Document Visual Question Answering using Self-Attention Scoring Mechanism}
\titlerunning{MP-DocVQA using Self-Attn Scoring}
%
\author{Lei Kang\inst{1}\orcidID{0000-0002-1962-3916} \and \\
Rubèn Tito\inst{1}\orcidID{0000-0002-5657-9790} \and \\
Ernest Valveny\inst{1}\orcidID{0000-0002-0368-9697} \and \\
Dimosthenis Karatzas\inst{1}\orcidID{0000-0001-8762-4454}}
\authorrunning{L. Kang et al.}
%
\institute{Computer Vision Center, Universitat Autònoma de Barcelona, Spain
\email{\{lkang,rperez,ernest,dimos\}@cvc.uab.es}\\
}
\maketitle              
\begin{abstract}
Documents are 2-dimensional carriers of written communication, and as such their interpretation requires a multi-modal approach where textual and visual information are efficiently combined. Document Visual Question Answering (Document VQA), due to this multi-modal nature, has garnered significant interest from both the document understanding and natural language processing communities. The state-of-the-art single-page Document VQA methods show impressive performance, yet in multi-page scenarios, these methods struggle. They have to concatenate all pages into one large page for processing, demanding substantial GPU resources, even for evaluation. In this work, we propose a novel method and efficient training strategy for multi-page Document VQA tasks. In particular, we employ a visual-only document representation, leveraging the encoder from a document understanding model, Pix2Struct. Our approach utilizes a self-attention scoring mechanism to generate relevance scores for each document page, enabling the retrieval of pertinent pages. This adaptation allows us to extend single-page Document VQA models to multi-page scenarios without constraints on the number of pages during evaluation, all with minimal demand for GPU resources. Our extensive experiments demonstrate not only achieving state-of-the-art performance without the need for Optical Character Recognition (OCR), but also sustained performance in scenarios extending to documents of nearly 800 pages compared to a maximum of 20 pages in the MP-DocVQA dataset. Our code is publicly available at \url{https://github.com/leitro/SelfAttnScoring-MPDocVQA}.

\end{abstract}
\section{Introduction}
Document Visual Question Answering (Document VQA~\footnote{In this paper, we term it "Document VQA" to avoid confusion with the DocVQA dataset.}) is a challenging problem that requires logical reasoning over text images mixed with typed and handwritten fonts, diverse layouts found in historical or modern documents, and complex components like diagrams, figures and tables. Moreover, documents like manuals or reports typically comprise multiple pages. Consequently, multi-page Document VQA has recently been proposed as a more practical challenge that could potentially find widespread use in real-world scenarios. In that sense, while the commonly used DocVQA~\cite{mathew2021docvqa} dataset consists of natural language questions on single-page text images, MP-DocVQA~\cite{tito2023hierarchical} dataset has been introduced as an extension of DocVQA to pose natural language questions on multi-page document images. 

Since Document VQA is a naturally multi-modal problem, a common approach is to transform the text content on the documents from visual modality into text using open-source or commercial Optical Character Recognition (OCR) tools. Nevertheless, this approach may pose significant constraints. On one hand, numerous document elements, including the layout, diagrams, figures, fonts, handwritten text, mathematical equations, and more, are required for reasoning together with the question in order to generate accurate answers. On the other hand, there is no guarantee that the text on the page is completely and accurately recognized by the OCR tools.

To address this limitation, most state-of-the-art approaches are proposed to utilize the visual content by either incorporating 2D positional information into the OCR tokens~\cite{powalski2021going}, or projecting image patches into the pre-trained language feature space~\cite{huang2022layoutlmv3}. Given the proven effectiveness of Transformers in recognizing both typed and handwritten fonts~\cite{kang2022pay}, an alternative approach would involve eliminating noisy OCR annotations entirely and harnessing the full potential of Transformers to extract textual information from documents. Thus, rich representations can be learned from the visual information, and subsequently used as a unified representation space for both the textual and visual modalities, as introduced in Pix2Struct~\cite{lee2023pix2struct}. This approach involves rendering of the textual information as images, and leveraging the model's inherent visual encoding capability to encode both the question and page context as a whole image. Thus, the answers can be generated by attending on question and page context features embedded in a unified visual feature space. 

While these methods excel in single-page Document VQA tasks, they encounter challenges when applied to multi-page scenarios. Concatenating all pages into a single large page for processing requires significant GPU resources, even during evaluation. Therefore, dedicated multi-page Document VQA methods are imperative. The current state-of-the-art method is Hi-VT5~\cite{tito2023hierarchical}, which is a multi-modal hierarchical encoder-decoder transformer based on T5~\cite{raffel2020exploring}. In its training, the input consists of questions, sequences of page images and the corresponding OCR annotations. 

In this paper, we introduce an OCR-free approach for the multi-page Document VQA task, leveraging the Pix2Struct document understanding model as our backbone. Our method is trained solely on imagery data, comprising the questions rendered as images and sequences of page images. We propose a Self-attention scoring mechanism to determine relevance scores for each document page based on a given question, facilitating the retrieval of relevant pages. This allows for minimal GPU resource usage when handling multi-page scenarios, without limitations on the number of pages during evaluation. The main contributions are summarized as follows:

\begin{itemize}
    \item We propose a self-attention scoring mechanism that adapts the single-page Document VQA method to the multi-page scenario with minimal GPU resource usage.
    \item We introduce an efficient training scheme where only one positive page and one randomly selected negative page of a document are incorporated for training the proposed scoring mechanism. 
    \item We present a new perspective to align textual and visual modalities by simply converting them into pixel-based representations.
    \item We conduct evaluation at a more realistic scenario by extending the MP-DocVQA dataset to include the complete documents with a maximum page count of $793$, as opposed to the original test set with a maximum of $20$ pages. Our method demonstrates satisfactory performance in this extended evaluation.
\end{itemize}

\section{Related Work}

As Document VQA involves processing two modalities of natural language questions and document images to generate natural language answers, it leads to the categorization of Document VQA methods into three main types of pipelines:

\textbf{Transforming visual modality into text.} A pre-processing using OCR annotation tools to transcribe the text content of document images into text is always the first step for this pipeline. Modern Handwritten Text Recognition (HTR) methods~\cite{coquenet2023dan,parres2023fine,jian2023hisdoc,kang2022pay} can also play a role in achieving reasonable recognition performance for handwritten documents. Then it becomes a conventional QA problem. In recent years, large pre-trained language models such as BERT~\cite{devlin2018bert}, GPT-3~\cite{brown2020language}, T5~\cite{raffel2020exploring}, Gopher~\cite{rae2021scaling} and Flan-T5~\cite{chung2022scaling} have achieved state-of-the-art results on public QA benchmarks. These methods can be employed for extracting answers from OCR annotations in document images, effectively addressing the Document VQA challenge. Nevertheless, there are two main disadvantages using OCR annotations as input. Firstly, there is no guarantee that they are completely accurate in terms of recognition with popular OCR annotation tools like Amazon Textract\footnote{https://aws.amazon.com/textract/}. Secondly, all non-textual information transmitted by the document is lost, including the 2D layout and reading order of the transcribed text.

\textbf{Aligning textual and visual modalities by learning a joint representation of text and images.}
Recently, in the domain of document understanding, there have been notable advancements in the form of highly capable multi-modal models based on large pre-trained transformers. Methods such as TILT~\cite{powalski2021going}, UDoc~\cite{gu2021unidoc}, DocFormer~\cite{appalaraju2021docformer}, SelfDoc~\cite{li2021selfdoc}, ERNIE-Layout~\cite{peng2022ernie} and LayoutLM-v3~\cite{huang2022layoutlmv3} have achieved state-of-the-art performance in Document VQA tasks. These methods may differ in their architectural designs and pre-training strategies, but they all adhere to a common concept: utilizing all accessible information from documents, including text and bounding boxes from OCR annotations, original image visual elements, and question text tokens as their input. These methods prove to be applicable in Document VQA tasks, yet their effectiveness hinges on the performance of OCR annotation tools when dealing with target documents. Donut~\cite{kim2022ocr} and Dessurt~\cite{davis2022end} are OCR-free methods for visual document understanding, by aligning text and visual attributes via a cross-attention mechanism. 

\textbf{Transforming textual modality into pixels.}
If we reverse our perspective, we can consider that all valuable information from document images, including text, layouts, figures, logos, signatures, tables, and more, is encoded within pixels. We may contemplate transforming natural language questions into pixels as well and using pixel-only input to train a Document VQA model. Pix2Struct~\cite{lee2023pix2struct} is a successful attempt of this approach that renders natural language text into TrueType font images, thereby transforming the input into a purely visual modality. It has been designed as a general document understanding model for various downstream tasks, including captioning and visual question answering. Pix2Struct offers a pre-trained model for single-page Document VQA, which leverages pre-training on HTML screenshots followed by fine-tuning on the DocVQA dataset~\cite{mathew2021docvqa}. But there remains a challenge to provide a definitive and effective way for extending its applicability to a multi-page scenario.

\textbf{Multi-page Document VQA methods.}
Document VQA in a multi-page scenario has not been explored sufficiently. Recently, the MP-DocVQA dataset~\cite{tito2023hierarchical} extended the original single page DocVQA dataset~\cite{mathew2021docvqa} for multi-page document VQA tasks. The MP-DocVQA dataset is constrained to have a maximum number of $20$ pages in each document. And all the questions end up focusing on the evidence in single pages, i.e., each answer can be obtained from a single page in the document targeting a given question. Additionally, the authors of this dataset introduced the Hi-VT5 model, which is a multi-modal hierarchical encoder-decoder method based on DiT~\cite{li2022dit} and T5~\cite{raffel2020exploring}. It accepts input of natural language questions, multiple page images, and their associated OCR annotations, with the goal of aligning these textual and visual modalities to create a unified representation. Nonetheless, it depends on OCR annotations, and the utilization of GPU memory escalates as the document page count grows, so that it would be incapable of handling documents with an extensive number of pages.

\section{Method}

The problem of multi-page Document VQA can be defined as follows: Given a combination of a question, a multi-page document, and the correct page identifier indicating which page to find the supporting evidence for the answer in, represented as $(Q_i, D_j, I_{ij})$, the goal is to predict the correct page identifier and generate a suitable answer as the result. Questions $\mathcal{Q} = \{Q_0, ..., Q_N\}$ are natural language texts. Documents $\mathcal{D} = \{D_0, ..., D_M\}$ contain a variable number of page images $D_j = \{p_0, ..., p_L\}$. The page identifier $I_{ij}$ takes values $\{0, 1\}$ where $1$ is the answer if the $Q_i$ question’s answer is present in Document $D_j$, and $0$ otherwise.

\begin{figure*}[h!]
    \centering
    \includegraphics[width=0.99\linewidth]{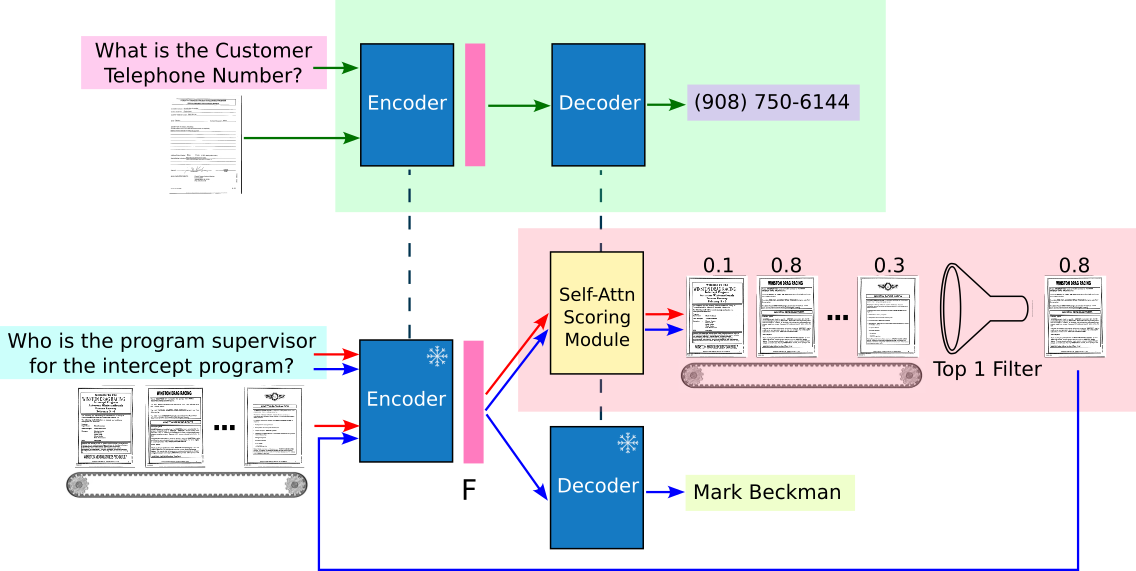}
    \caption{Overview of the proposed framework. 
The training process involves two steps: initially training the single-page model in the upper green block with positive question-page pairs only, then freezing the single-page model and training the self-attention scoring module in the red block with positive and negative question-page pairs to retrieve the most relevant document pages. The evaluation is indicated by blue arrows.}
    \label{fig:arch}
    \vspace{-0.4cm}
\end{figure*}

Our proposed method follows a two-stage training scheme as shown in green and red arrows of Fig.~\ref{fig:arch}. The first stage, illustrated by green arrows, trains a Document VQA model in single page scenario using input pairs $(Q_i, p_k)$, where $p_k$ is the correct page of document $D_j$ when $I_{ij}$ is $1$. Once the single-page Document VQA model is properly trained, the model weights will be frozen for their usage in the second stage. The second stage, depicted by red arrows, utilizes the contextual feature $F$ extracted from the frozen Encoder in the first stage, producing a matching score between $0$ and $1$. The higher the matching score value, the more likely the question is related to the page. The evaluation is firstly conducted by passing each page of the target document with the question iteratively into the Encoder and self-attention scoring module to obtain the matching scores for each question-page pair. Then, a top-1 filter is applied to select the most related question-page pair according to the produced probability. Finally, the chosen question-page pair goes into the Encoder and Decoder, generating the answer. The evaluation pipeline is illustrated in blue arrows of Fig~\ref{fig:arch}. 

\begin{figure}[h!]
    \centering
    \includegraphics[width=0.8\linewidth]{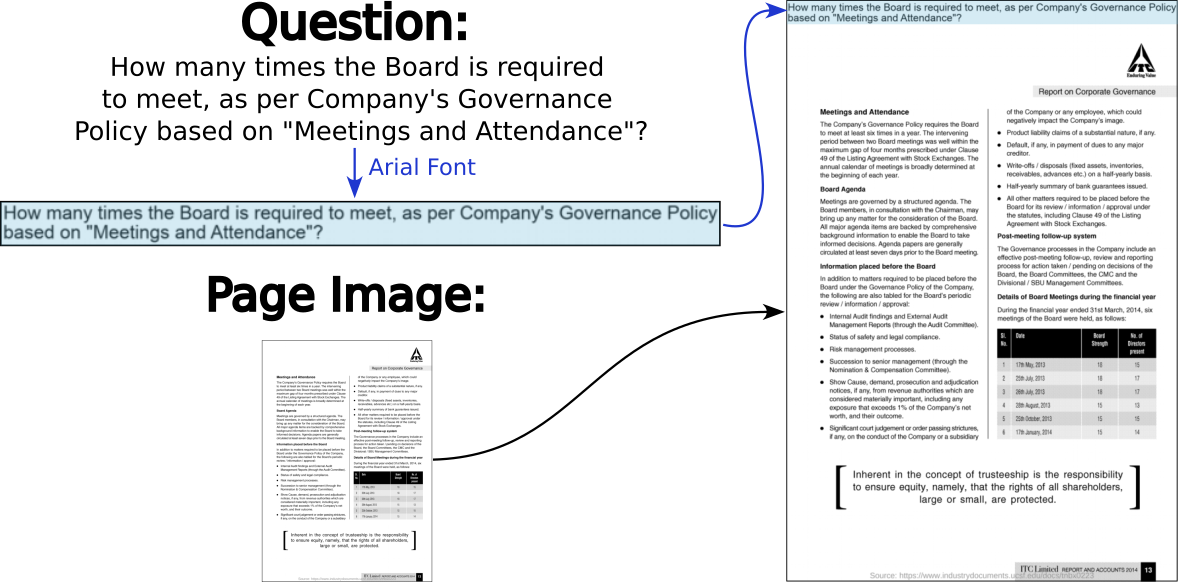}
    \caption{Transforming natural language questions into visual modality. Note that the blue color solely serves for the purpose of highlighting the text within the image, which is not integrated into the final concatenated image.}
    \label{fig:text2pixel}
    \vspace{-0.2cm}
\end{figure}

\vspace{-0.6cm}

\subsection{Single-page Document VQA Model}
 We choose Pix2Struct-Base~\cite{lee2023pix2struct} as our baseline model. We were able to bypass the first-stage training phase thanks to the Hugging Face open-source community~\footnote{https://huggingface.co/google/pix2struct-docvqa-base}. We utilize the pre-trained weights they provided from fine-tuning on the DocVQA dataset~\cite{mathew2021docvqa}. The model has an encoder-decoder architecture based on transformers. The input question $Q_i$ undergoes a preliminary transformation into an image employing a TrueType font. Subsequently, the question image is concatenated to the top of the page $p_k$, resulting in an extended image with both question and page information in visual format. This process is illustrated in Fig.~\ref{fig:text2pixel}. This extended image is further adjusted in size while maintaining its aspect ratio to ensure compatibility for fitting into a maximum length of 2048 patches with dimensions of $16\times16$. This training pipeline is indicated with \textcolor{green}{Green Arrows} in Fig.~\ref{fig:arch}. The Encoder feature $F$ is obtained by processing the question and page image patches. After the Encoder-Decoder architecture is properly trained on the single-page Document VQA task, the Encoder feature $F$ is expected to acquire a good contextual representation between the question and the page. The Encoder feature $F$ is to be utilized for the second stage training.

\subsection{Self-attention Scoring Module}

\begin{figure}[h!]
    \centering
    \includegraphics[width=0.8\linewidth]{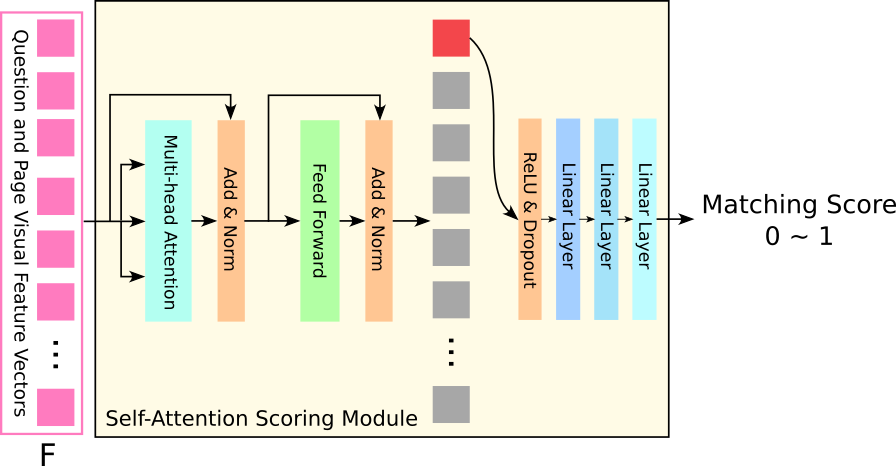}
    \caption{The architecture of Self-Attention Scoring Module.}
    \label{fig:scoring}
    \vspace{-0.4cm}
\end{figure}

We propose a self-attention scoring module to evaluate how good is the match between the question and the page as detailed in Fig.~\ref{fig:scoring}. The self-attention scoring module consists of a self-attention layer and three linear layers with a dropout layer in between. The input is the encoded feature $F$, which is a sequence of question and page visual feature vectors. The self-attention layer's output shares the same length as the $F$ feature vectors. Only the first vector is selected serving as the matching representation, as indicated as red rectangle of Fig.~\ref{fig:scoring}. First-vector-method is utilized according to experimental ablation study in comparison with other aggregation methods detailed in Sec.~\ref{sec:abla}. Then, a matching score between $0$ and $1$ is produced by passing the representation through a dropout layer and three linear layers.


During training, we consider the pages containing the evidence for the answers as positive pages, and the other pages as negative ones. The positive/negative labels are utilized to train the self-attention scoring module. In the ground-truth training set, there is only one positive page $p_k$ in the document $D_j$ targeting the question $Q_i$. Thus, there are a lot more negative pages than the positive ones in the multi-page scenario. To balance the training data, only one negative page $p_r$ is randomly selected for each question-document input $(Q_i, D_j)$. In other words, one positive page question pair $(Q_i, p_k)$ is paired with one randomly selected negative page question pair $(Q_i, p_r)$. In the phase of input data preprocessing, the question $Q_i$ is rendered into a text image, and subsequently concatenated to the top of both the positive page $p_k$ and negative page $p_r$ in the document $D_j$. The encoded feature $F$, obtained from the Encoder with frozen weights, is fed into the self-attention scoring module, resulting in a matching score that represents a probability from $0$ to $1$. The random negative data selection strategy enhances training efficiency by excluding certain pages of documents from training. This training pipeline is indicated with \textcolor{red}{Red Arrows} in Fig.~\ref{fig:arch}. The training is performed utilizing the Mean Squared Error (MSE) loss while applying label smoothing. More details can be found in our code.

\subsection{Evaluation Process}
During testing, every page $p_t$ of the document $D_j$ is processed with question $Q_i$ as input to the Encoder to obtain the question-page pair contextual feature $F$. The scoring module takes the encoded feature $F$ as input and generates a matching score for the given question-page pair $(Q_i, p_t)$. Once this process has been done for all the pages of the target document, a straightforward Top-1 filtering module is employed to choose the question-page pair with the highest probability. Thereafter, the full Encoder-Decoder pipeline is applied with the selected single page with the question, generating the desired answer. The final evaluation pipeline is indicated with \textcolor{blue}{Blue Arrows} in Fig.~\ref{fig:arch}.

\section{Experiments}

\subsection{Dataset and Metrics}
\begin{figure}[h!]
    \centering
    \includegraphics[width=0.99\linewidth]{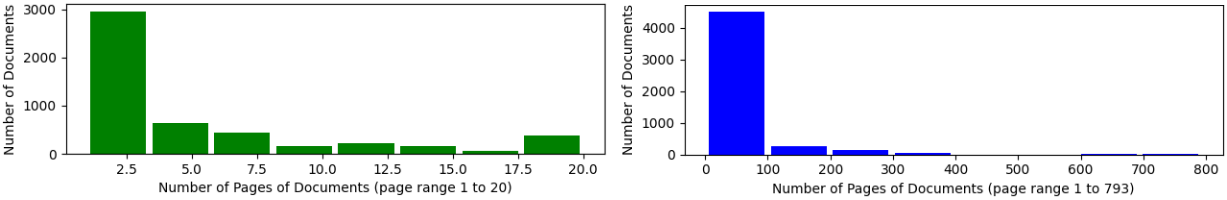}
    \caption{Histograms of the number of pages in the original test set documents and the extended version with full pages, depicting in green and blue charts, respectively.}
    \label{fig:histo}
    \vspace{-0.4cm}
\end{figure}

The recently released MP-DocVQA dataset~\cite{tito2023hierarchical} targets the Document VQA problem in a multi-page scenario. It comprises $46,000$ questions across $6,000$ documents, each with a variable number of pages artificially limited to $20$. Note that in the ground-truth of the dataset, the correct answer is contained within a single page for a particular question-document combination. To assess the robustness and practicality of our method in a more challenging and realistic scenario, we propose to remove the restrictions on the page length of the documents. This extension allows for documents of up to 800 pages. The increase in the number of pages is not implemented on the training set, so that we can maintain a fair comparison with the state of the art without introducing additional information during training. The histograms of the number of pages for both the original and extended test set are shown in Fig.~\ref{fig:histo}.

\vspace{-0.4cm}

To ensure a fair comparison, our work employs the standardized evaluation metrics, namely, Average Normalized Levenshtein Similarity (ANLS), as originally introduced along the DocVQA dataset~\cite{mathew2021docvqa}. To ensure accurate page prediction, we employ accuracy (\%) as the metric.

\subsection{Hyper-parameter Search and Ablation Study}
\label{sec:abla}
\begin{figure}[!htb]
\minipage{0.49\textwidth}
\vspace{0.4cm}
    \includegraphics[width=0.99\linewidth]{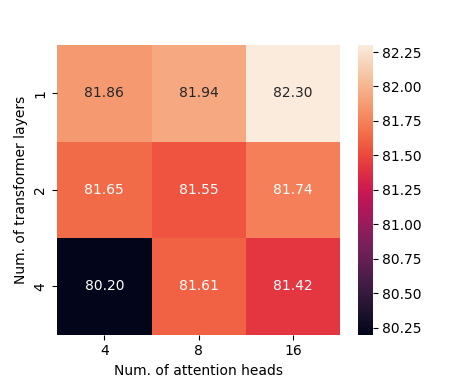}
    \caption{Heatmap of page prediction accuracy ($\%$) on validation set.}
    \label{fig:heatmap_page}
\endminipage
\minipage{0.49\textwidth}
    \includegraphics[width=0.99\linewidth]{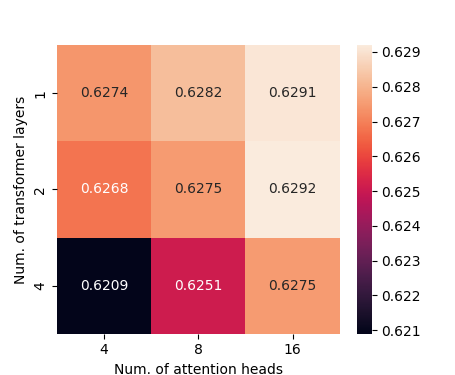}
    \caption{Heatmap of ANLS on validation set.}
    \label{fig:heatmap_anls}
\endminipage
\end{figure}

As the Pix2Struct-Base~\cite{lee2023pix2struct} is utilized as our backbone model, we keep the same hyper-parameters as in the original paper. For the implementation of the proposed self-attention scoring module, we conduct a hyper-parameter search to identify an optimal combination of the number of self-attention layers and attention heads, evaluating with both page prediction accuracy and ANLS, as shown in Fig.~\ref{fig:heatmap_page} and Fig.~\ref{fig:heatmap_anls}, respectively. 



The figures show that the best accuracy for page prediction on the validation set is obtained when using a configuration with a single self-attention layer and $16$ attention heads, as seen in Figure~\ref{fig:heatmap_page}. Similarly, when examining the ANLS performance in Figure~\ref{fig:heatmap_anls}, we observe that optimal results are achieved with either one or two self-attention layers and $16$ attention heads. In order to maintain the simplicity of the proposed model, we opt for a configuration with one self-attention layer and $16$ attention heads.

In the self-attention scoring module depicted in Fig.~\ref{fig:scoring}, there exists an aggregation method aimed at distilling a sequence of vectors into a fixed-size vector. This process is visually represented as a sequence of squares colored red and gray. We propose three methods: firstly, by introducing a special $[CLS]$ token at the beginning of the input feature vectors, we aim to signal the scoring module to aggregate global features into the initial output vector. Secondly, by utilizing adaptive average pooling, we are also able to aggregate the variable-length feature vectors into a fixed-size vector. Lastly, by simply selecting the first vector, the model can be trained to aggregate global information into a fixed-size vector. The ablation study on the three aggregation methods is shown in Table~\ref{tab:aggre}. Based on the results in the Table, selecting first vector is the best option to aggregate global information from the sequence of vectors. Incorporating a $[CLS]$ token to explicitly force model aggregate global information to the initial vector performs the worst because this involves an extra alignment process between special token embedding and input visual feature vectors. Adaptive average pooling has achieved reasonable performance. However, because the pooling-out process is not guided by the training goal, randomization may potentially hinder performance. The proposed method of selecting the first output vector of the encoded feature has achieved the best performance. During the training process, we enforce the aggregation of global information into the first output vector, which proves to be both straightforward and effective.

\begin{table*}[h!]
    \caption{Ablation study on aggregation methods inside self-attention scoring module.}
    \vspace{-0.1cm}
    \label{tab:aggre}
    \centering
    \small
    \scalebox{0.99}{
    \begin{tabular}{cc}
    \toprule
    \textbf{Method} & \textbf{Page Pred. (\%)} \\ 
    \midrule
    $[CLS]$ token & 60.56\\
    AdaptAvgPool & 81.95\\
    \textbf{First Vector (proposed)} & \textbf{82.19}\\
    \bottomrule
    \end{tabular}
    }
    \vspace{-0.2cm}
\end{table*}

\subsection{Comparison with the State of the Art}
Using the optimal hyper-parameter settings, we train the self-attention scoring module on the training set, employing early-stopping after 5 epochs on the validation set. The final results by evaluating the proposed method on the test set are shown in the last row of Tab.~\ref{tab:sota}. Our proposed method outperforms the state-of-the-art methods in page prediction thanks to the simple yet efficient self-attention scoring module. The ANLS performance of the proposed method also ranks favorably among the current state-of-the-art methods. Comparing with the current state-of-the-art method Hi-VT5~\cite{tito2023hierarchical}, the proposed method achieves similar performance without the need for OCR annotations and with fewer parameters. Additionally, our proposed method has the capability to handle lengthy documents without page limitations.

\begin{table*}[h!]
    \caption{Comparison with state of the art on the test set of MP-DocVQA dataset.}
    \vspace{-0.1cm}
    \label{tab:sota}
    \centering
    \small
    \scalebox{0.99}{
    \begin{tabular}{cccccc}
    \toprule
    \textbf{Method} & \textbf{Year} & \textbf{OCR} & \textbf{Params.} & \textbf{Page Pred. (\%)} & \textbf{ANLS}\\ 
    \midrule
    BERT~\cite{devlin2018bert} & 2018 & \checkmark & 334M & 71.24 & 0.5347\\
    T5~\cite{raffel2020exploring} & 2020 & \checkmark & 223M & 46.05 & 0.4028\\
    Longformer~\cite{beltagy2020longformer} & 2020 & \checkmark & 148M & 70.37 & 0.5506\\
    Big Bird~\cite{zaheer2020big} & 2020 & \checkmark & 131M & 72.27 & 0.5854\\
    LayoutLMv3~\cite{huang2022layoutlmv3} & 2022 & \checkmark & 125M & 74.02 & 0.5513\\
    Hi-VT5~\cite{tito2023hierarchical} & 2023 & \checkmark & 316M & 79.23 & \textbf{0.6201}\\
    \midrule
    \textbf{proposed} & 2024 & \textbf{$-$} & 273M & \textbf{81.55} & 0.6199\\
    \bottomrule
    \end{tabular}
    }
    \vspace{-0.2cm}
\end{table*}


\subsection{Performance in Unrestricted Scenario}
The released MP-DocVQA dataset has the artificial constraint that the maximum number of pages in each document is limited to $20$, which is already a challenging scenario. But in our experiments, we would like to evaluate the proposed method in the hardest condition, where the complete multi-page documents are available without any limitation on the number of pages. For both the validation and test sets, we extend the number of pages of each document to the complete original number of pages. The results are illustrated in Tab.~\ref{tab:extreme}. From the Table we can see that the maximum number of pages of documents has reached $613$ and $793$ for validation and test sets, respectively. The average number of pages in both the validation and test sets is $7$ times greater than the original average. The total number of pages has increased roughly 7-fold compared to the original data. Nonetheless, the page prediction performance experiences decreases by approximately $25\%$, while the ANLS performance declines by less than $13\%$.

\begin{table*}[!th]
    \caption{Performance comparison between original set and extended full-page version for both validation and test sets.}
    \vspace{-0.1cm}
    \label{tab:extreme}
    \centering
    \small
    \scalebox{0.98}{
    \begin{tabular}{ccccccc}
    \toprule
    \multirow{2}{*}{\textbf{Set}} & \multicolumn{4}{c}{\textbf{Number of Pages}} & \multicolumn{1}{c}{\textbf{Page Pred.}} & \multirow{2}{*}{\textbf{ANLS}}\\
    & Min. & Avg. & Max. & Total & \textbf{(\%)} & \\
    \midrule
    Valid-Original & 1 & 5.6 & 20 & 28,906 & 81.26 & 0.6252\\
    Valid-Extended & 1 & 37.2 & 613 & 192,775 & 61.15 & 0.5460\\
    \midrule
    Test-Original & 1 & 5.1 & 20 & 25,630 & 81.55 & 0.6199\\
    Text-Extended & 1 & 38.5 & 793 & 193,216 & 60.45 & 0.5394\\ 
    \bottomrule
    \end{tabular}
    }
    \vspace{-0.2cm}
\end{table*}
\vspace{-0.4cm}

\subsection{Thorough Analysis on Validation Set}
While we are able to access the entire set of unlabeled pages in the dataset to expand the validation and test sets to include complete pages, it is important to note that the ground-truth data, including the correct answers and page identifiers, for the test set is not publicly accessible. As a result, we obtain the evaluation results for the test set through the official competition website~\footnote{https://rrc.cvc.uab.es/?ch=17}. Hence, we conduct a thorough analysis on the validation set instead. In Tab.~\ref{tab:analysis}, we analyze the performance correlation between page prediction and ANLS for question-document pairs. For both the original and extended versions of the validation sets, an interesting observation can be made by analysing the results in third rows, while the error is made in page prediction but ANLS is fully correct. It is worth analyzing why this would happen.
\vspace{-0.2cm}
\begin{table}[t!h]
    \caption{Statistics on the original and extended validation data.}
    \vspace{-0.1cm}
    \label{tab:analysis}
    \centering
    \small
    \scalebox{0.85}{
    \begin{tabular}{ccccc}
    \toprule
    \textbf{Valid. Set}  & \textbf{Corr. Page Pred.} & \textbf{ANLS==1} & \textbf{Count} & \textbf{Pct. (\%)}\\ 
    \midrule
    \multirow{4}{*}{Original} & \checkmark & \checkmark & 2488 & 47.97\\
    & \checkmark & \xmark & 1727 & 33.29 \\
    & \xmark & \checkmark & 172 & 3.32\\
    & \xmark & \xmark & 800 & 15.42\\
    \midrule
    \multirow{4}{*}{Extended} & \checkmark & \checkmark & 1928 & 37.17\\
    & \checkmark & \xmark & 1244 & 23.98\\
    & \xmark & \checkmark & 357 & 6.88\\
    & \xmark & \xmark & 1658 & 31.96\\
    \bottomrule
    \end{tabular}
    }
    \vspace{-0.2cm}
\end{table}




\begin{figure}[h!]
\minipage{\textwidth}
    \includegraphics[width=0.9\linewidth]{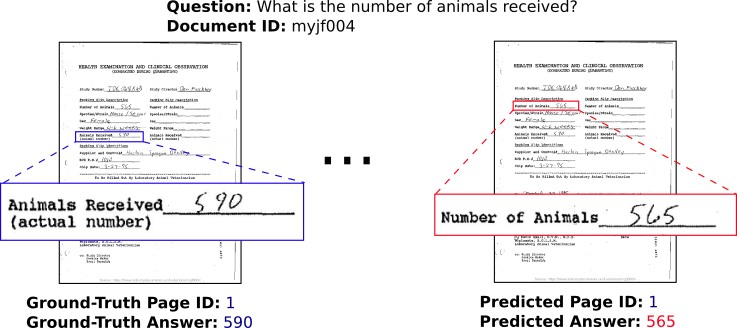}
    \caption{A failure case with the correct page prediction and failed answer in the validation set, corresponding to the 2nd row of Tab.~\ref{tab:analysis}.}
    \label{fig:case_10}
    \endminipage
    \\

\minipage{\textwidth}
\vspace{0.4cm}
    \includegraphics[width=0.9\linewidth]{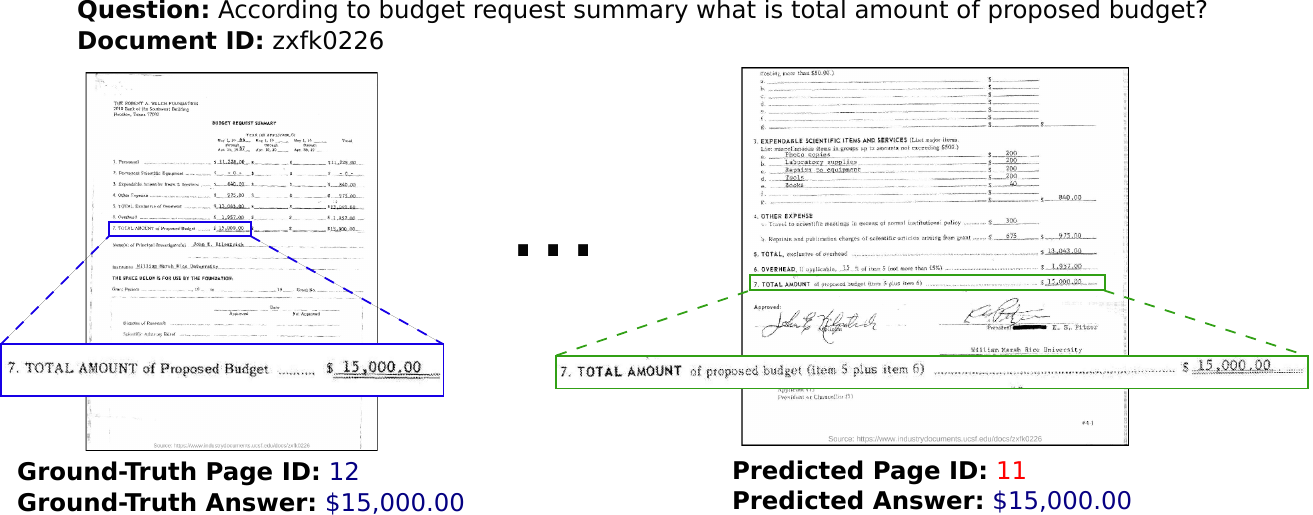}
    \caption{A failure case with the failed page prediction and correct answer in the validation set, corresponding to the 3rd row of Tab.~\ref{tab:analysis}.}
    \label{fig:case_01}
\endminipage
\\
\minipage{\textwidth}
\vspace{0.4cm}
    \includegraphics[width=0.9\linewidth]{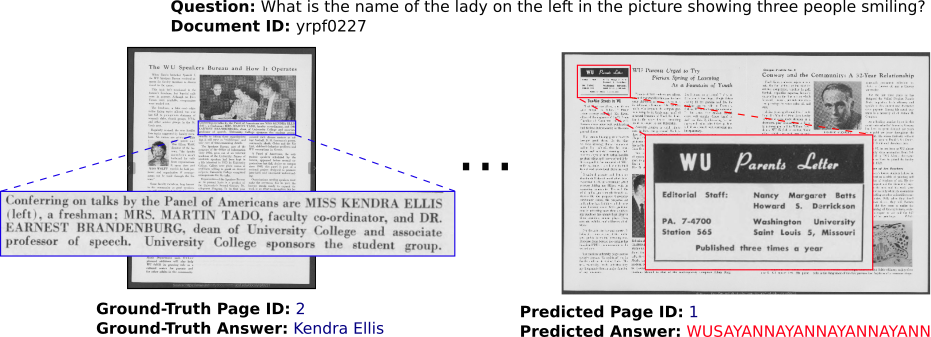}
    \caption{A failure case with the failed page prediction and failed answer in the validation set, corresponding to the 4th row of Tab.~\ref{tab:analysis}.}
    \label{fig:case_00}
\endminipage
\end{figure}

We have presented failure cases as shown in Fig.~\ref{fig:case_10}, Fig.~\ref{fig:case_01} and Fig.~\ref{fig:case_00}, corresponding to the cases in the 2nd, 3rd and 4th row in Tab.~\ref{tab:analysis}, respectively. In Fig.~\ref{fig:case_10}, it is evident that the page prediction is correct but the answer is incorrect. This is due to that the supporting evidence (shown as bounding-boxes) has been located incorrectly. In Fig.~\ref{fig:case_01}, it showcases the interesting case that the page prediction is incorrect yet the answer is correct. We can see that the supporting evidence on both the 11th and 12th pages is sufficiently accurate to yield the desired answer. One of the assumptions made for the MP-DocVQA dataset is that in each document, only one page holds the evidence necessary to generate the answer. However, this assumption is not true, especially in the extended version with additional pages. In Fig.~\ref{fig:case_00}, both the page prediction and the answer are incorrect. This example demonstrates that answering the question necessitates the ability to reason about the figures on the page and understand the associated captions, which is a challenging task. In our prediction, it seems that the model prioritizes the "left" and "picture" over the finer details of "lady" and "three people smiling".

From these cases, we are aware that the evidence required to infer the answers is provided on more than one page in the document, especially in the extended version including the full document pages. Thus, it is reasonable, as illustrated in Tab.~\ref{tab:analysis}, to obtain $3.32\%$ and $6.88\%$ of data with failed page prediction yet correct answers in the original and extended validation sets, respectively.

\section{Conclusion}
In this paper, we present a multi-page Document VQA method that employs a novel self-attention scoring mechanism. Our proposed method stands out for its simplicity and efficiency, achieved through a 2-stage training process involving a single-page Document VQA model and a scoring module. Notably, our method excels in handling real-world scenarios involving target documents of varying page counts, without imposing any page limitations. 


\section*{Acknowledgements}
Beatriu de Pinós del Departament de Recerca i Universitats de la Generalitat de Catalunya (2022 BP 00256), European Lighthouse on Safe and Secure AI (ELSA) from the European Union’s Horizon Europe programme under grant agreement No 101070617.


\bibliographystyle{splncs04}
\bibliography{custom.bib}
\end{document}